\documentclass[10pt,twocolumn,letterpaper]{article}

\usepackage[pagenumbers]{cvpr} %

\usepackage[table]{xcolor}

\usepackage{multirow}

\usepackage{colortbl}

\usepackage{booktabs}

\usepackage{bm}

\usepackage{graphicx}

\usepackage{caption}

\usepackage{array}

\usepackage{amssymb}

\usepackage{amsmath}

\usepackage{wrapfig}

\usepackage{booktabs,makecell,multirow,pifont}

\usepackage{marvosym}
\newcommand{\corrmark}{\begingroup\renewcommand\thefootnote{\Letter}\footnotemark\endgroup}

\newcommand{\modelname}{GPOcc}

\definecolor{cvprblue}{rgb}{0.21,0.49,0.74}
\usepackage[pagebackref,breaklinks,colorlinks,allcolors=cvprblue]{hyperref}

\def\paperID{} %
\def\confName{CVPR}
\def\confYear{2026}

\title{Generalizing Visual Geometry Priors to Sparse Gaussian Occupancy Prediction}

\author{
Changqing Zhou$^{1}$ \quad
Yueru Luo$^{2}$ \quad
Changhao Chen$^{1}$\corrmark
\vspace{3pt} \\
$^{1}$ The Hong Kong University of Science and Technology (Guangzhou) \\
$^{2}$The Chinese University of Hong Kong, Shenzhen \\
{\tt\small czhou149@connect.hkust-gz.edu.cn \quad changhaochen@hkust-gz.edu.cn}
}

\begin{document}
\maketitle
\begingroup\renewcommand\thefootnote{\Letter}\footnotetext{Corresponding author.}\endgroup

\definecolor{ceiling}{RGB}{214,  38, 40}   %
\definecolor{floor}{RGB}{43, 160, 4}     %
\definecolor{wall}{RGB}{158, 216, 229}  %
\definecolor{window}{RGB}{114, 158, 206}  %
\definecolor{chair}{RGB}{204, 204, 91}   %
\definecolor{bed}{RGB}{255, 186, 119}  %
\definecolor{sofa}{RGB}{147, 102, 188}  %
\definecolor{table}{RGB}{30, 119, 181}   %
\definecolor{tvs}{RGB}{160, 188, 33}   %
\definecolor{furniture}{RGB}{255, 127, 12}  %
\definecolor{objects}{RGB}{196, 175, 214} %

\begin{figure*}[t]
    \centering
    \scriptsize
    \newcolumntype{P}[1]{>{\centering\arraybackslash}p{#1}}
    \setlength{\tabcolsep}{0.01\linewidth} %
    \renewcommand{\arraystretch}{1.5}      %
    \begin{tabular}{P{0.31\textwidth} P{0.31\textwidth} P{0.31\textwidth}}
        \includegraphics[width=0.95\linewidth]{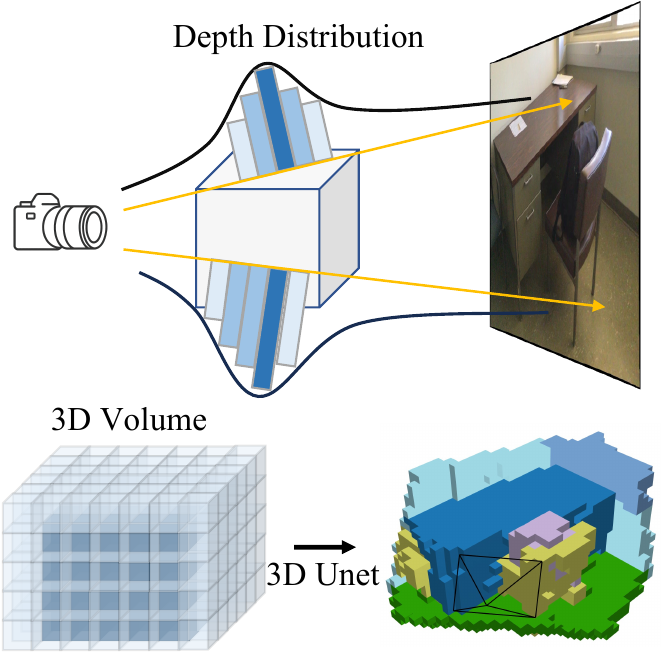} &  
        \includegraphics[width=0.95\linewidth]{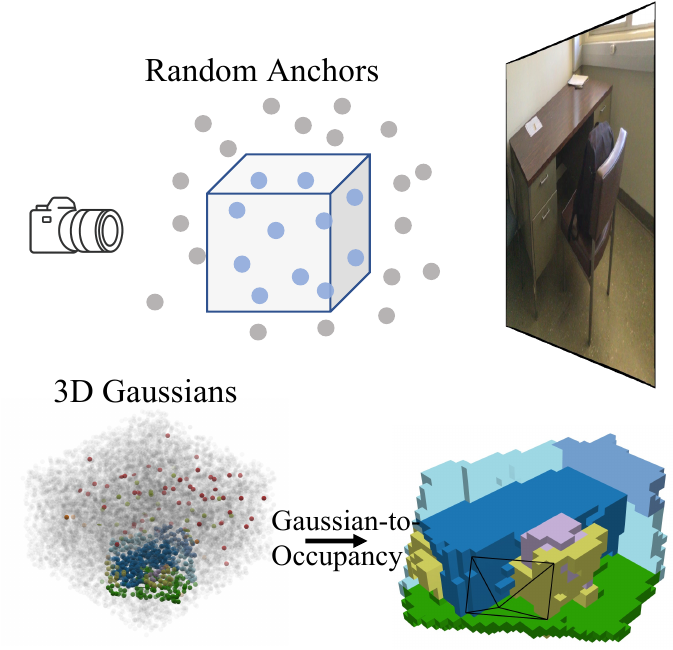} &  
        \includegraphics[width=0.95\linewidth]{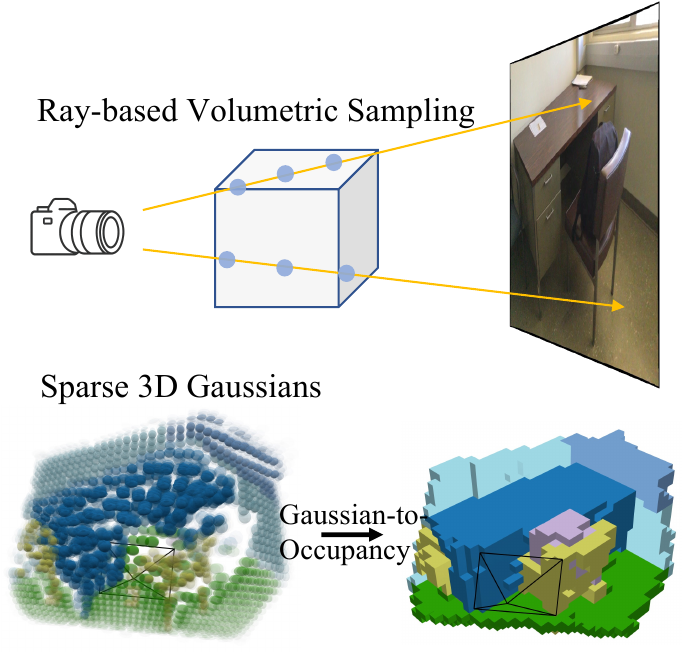} \\
        (a) ISO~\cite{ISO} & (b) EmbodiedOcc~\cite{embodiedocc} & (c) Ours
    \end{tabular}
\caption{\textbf{Comparison of monocular occupancy prediction pipelines.} 
ISO~\cite{ISO} formulates depth estimation as a multi-class classification problem, using the predicted depth distributions to lift 2D image features into dense 3D volumes, which are then processed by a 3D U-Net for occupancy prediction. 
EmbodiedOcc~\cite{embodiedocc}, by contrast, initializes random 3D anchors and applies cross-attention to aggregate image features, predicting Gaussian primitives that are splatted into voxels. Many of these Gaussians fall in empty regions, shown as gray primitives. 
In contrast, \modelname~employs ray-based volumetric sampling to generate sparse Gaussians concentrated on or within objects, producing a compact and efficient representation for occupancy inference.}
    \label{fig:teaser}
    \vspace{-3mm}
\end{figure*}

\begin{abstract}

Accurate 3D scene understanding is essential for embodied intelligence, with occupancy prediction emerging as a key task for reasoning about both objects and free space. Existing approaches largely rely on depth priors (e.g., DepthAnything) but make only limited use of 3D cues, restricting performance and generalization.
Recently, visual geometry models such as VGGT have shown strong capability in providing rich 3D priors, but similar to monocular depth foundation models, they still operate at the level of visible surfaces rather than volumetric interiors, motivating us to explore how to more effectively leverage these increasingly powerful geometry priors for 3D occupancy prediction.
We present \modelname, a framework that leverages generalizable visual geometry priors (GPs) for monocular occupancy prediction. Our method extends surface points inward along camera rays to generate volumetric samples, which are represented as Gaussian primitives for probabilistic occupancy inference. To handle streaming input, we further design a training-free incremental update strategy that fuses per-frame Gaussians into a unified global representation.
Experiments on Occ-ScanNet and EmbodiedOcc-ScanNet demonstrate significant gains: \modelname~improves mIoU by +9.99 in the monocular setting and +11.79 in the streaming setting over prior state of the art. Under the same depth prior, it achieves +6.73 mIoU while running 2.65$\times$ faster. These results highlight that \modelname~leverages geometry priors more effectively and efficiently.
Code will be released at \url{https://github.com/JuIvyy/GPOcc}.
\end{abstract}
    
\section{Introduction}

Embodied AI agents are increasingly expected to acquire accurate and detailed 3D understanding of their surroundings~\cite{embodiedaisurvey2025}, which is fundamental for reasoning, planning, and interaction in complex environments. In this process, vision plays a central role by providing rich semantic and geometric cues, and recent progress in vision-based 3D scene understanding has been substantial~\cite{bevformer,unimode,indoordepth,imvoxelnet,occfor3ddet}. Among various scene representations, occupancy prediction~\cite{ISO,embodiedocc} has gained particular traction by providing a unified and flexible volumetric model of both foreground objects and background structures, serving as a core building block for downstream tasks such as robotic navigation~\cite{volumetrivln}, interactive manipulation, and autonomous driving~\cite{occvla,occllama}.

While vision-centric occupancy prediction has been extensively studied in autonomous driving~\cite{voxformer,surroundocc,Triformer,occupancypoints,opus,gaussianformer,gaussianformer2}, fine-grained occupancy prediction in indoor scenarios remains considerably more challenging and less explored. The difficulty arises from cluttered spatial layouts and the wide diversity of object categories. Recent methods such as ISO~\cite{ISO} and EmbodiedOcc~\cite{embodiedocc} have made progress by incorporating depth priors~\cite{depthanything,depthanythingv2}.
For instance, ISO lifts 2D image features into dense 3D volumes using estimated depth distribution, as shown in~\Cref{fig:teaser}(a). The volumetric features are then processed with a 3D U-Net to predict the final occupancy.
In contrast, EmbodiedOcc initializes Gaussian primitives randomly and refines them through iterative cross-attention with image features. By projecting 3D anchors into images and subsequently splatting the refined Gaussians into occupancy prediction, as illustrated in~\Cref{fig:teaser}(b).
Although effective, these approaches make only limited use of depth priors and incur substantial redundancy by representing vast empty regions, which ultimately constrains performance and generalization.

Concurrently, a family of \emph{visual geometry priors} has emerged, ranging from monocular depth foundation models such as DepthAnything family~\cite{depthanything,depthanythingv2} to multi-view Visual Geometry Models (VGMs)~\cite{vggt,dust3r,mast3r,fast3r,pi3,point3r}. 
These models capture rich scene attributes (depth, point maps, and camera parameters) and enable high-quality 3D reconstruction, thereby providing strong geometric cues that could, in principle, greatly benefit occupancy prediction.
However, their outputs are inherently surface-centric: depth and point maps are restricted to visible surfaces, with each pixel typically corresponding to a single 3D surface point.
As a result, volumetric interiors remain unrepresented, and directly deriving occupancy from such geometry priors is non-trivial, often resulting in limited accuracy.

To bridge this gap, we introduce \textbf{\modelname}, a novel framework that leverages generalizable \textbf{G}eometry \textbf{P}riors (GPs) for \textbf{Occ}upancy prediction with sparse Gaussians rendering. Our approach builds on four key components:
\textbf{(1)} To overcome the limitation that GPs provide only surface points, we design a \textit{ray-based volumetric sampling} module that extends points inward along their camera rays, as illustrated in~\Cref{fig:teaser}(c). Each extended point predicts a Gaussian primitive that captures its local region.
\textbf{(2)} We adopt an \textit{opacity-based pruning} strategy that discards low-opacity Gaussians, significantly reducing redundancy with negligible performance loss (\Cref{tab:abl_opa}).
\textbf{(3)} Occupancy is inferred from the remaining sparse Gaussians using a probabilistic formulation following~\cite{gaussianformer2}.
\textbf{(4)} To adapt the framework to embodied scenarios with streaming input, we develop a training-free \textit{incremental update strategy} that incrementally fuses per-frame Gaussians into a coherent global representation.

We evaluate \modelname~on both the monocular Occ-ScanNet~\cite{ISO} dataset and the streaming EmbodiedOcc-ScanNet~\cite{embodiedocc} benchmark. With VGGT~\cite{vggt} as the prior, our method surpasses the previous state of the art by 9.99 mIoU and 8.24 IoU in the monocular setting, and by 11.79 mIoU and 9.21 IoU in the streaming setting. Moreover, when using the same depth prior~\cite{depthanythingv2} as EmbodiedOcc~\cite{embodiedocc}, \modelname~achieves gains of 6.73 mIoU and 3.41 IoU, while running at 2.65$\times$ FPS.
These results demonstrate both the accuracy and efficiency of our approach, validating the effectiveness of our novel application of general GPs in 3D occupancy.

We summarize our contributions as follows:

\begin{enumerate}[leftmargin=5.8mm]
\item We propose \modelname, a novel framework for 3D occupancy prediction that combines geometry priors with sparse continuous Gaussians, enabling fine-grained volumetric prediction in challenging indoor scenarios.
\item To address the limitation that geometry foundation model predict only visible surfaces, we introduce a ray-based volumetric sampling strategy that effectively reconstructs volumetric interiors from surface-based geometry priors.
\item We present a sparse Gaussian to occupancy formulation with opacity-based pruning and a training-free incremental update strategy, which together improve efficiency and extend the model to streaming video inputs.
\item We conduct extensive experiments and ablations, showing that \modelname~consistently achieves state-of-the-art performance on public datasets and generalizes effectively across different 3D geometry priors.
\end{enumerate}

\section{Related Works}

\subsection{Visual Geometry Foundation Model}

Early visual geometry models such as DUSt3R~\cite{dust3r} and MASt3R~\cite{mast3r} predict coupled scene representations (e.g., camera poses and geometry parameterized by pointmaps) from image pairs, but require expensive post-processing and symmetric inference for unconstrained multi-view SfM. Subsequent works including Spann3R~\cite{spann3r}, CUT3R~\cite{cut3r}, and MUSt3R~\cite{must3r} eliminate the reliance on classical optimization by introducing latent state memory in transformers, enabling multi-view reconstruction in a more end-to-end manner. Fast3R~\cite{fast3r} further scales this paradigm to handle over 1000 input images efficiently.
Building on this line, VGGT~\cite{vggt} employs an attention transformer to jointly predict pointmaps, depth, poses, and tracking features, relying on minimal 3D inductive biases but leveraging large-scale training data. StreamVGGT~\cite{streamvggt} reformulates VGGT with a causal transformer to overcome memory explosion when processing long video sequences. Several variants extend VGGT: $\pi^3$~\cite{pi3} fine-tunes VGGT to remove the reliance on the first input frame as the reference coordinate system, while Dens3R~\cite{dens3r} introduces normal prediction to enrich geometric cues.

\subsection{Occupancy Prediction}

MonoScene~\cite{monoscene} propelling semantic scene completion~\cite{monoscene,ndcscene,sscnet} into more challenging 3D occupancy with only image as input. While occupancy prediction has been extensively studied in outdoor autonomous driving scenarios~\cite{sparseocc,gaussianformer,gaussianformer2,voxformer}, research on indoor settings remains relatively limited~\cite{ISO,embodiedocc,embodiedocc++}.
Existing methods adopt different representations and strategies. Some approaches employ fixed 3D grids or voxel representations, where 2D image features are lifted into dense volumetric grids using depth distributions or predictions along camera rays, followed by 3D convolution or volumetric decoders~\cite{monoscene,ndcscene,ISO,fbocc,lss}. Transformers have also been explored for volumetric representation learning~\cite{occupancypoints}. Other methods rely explicitly on depth information or signed distance embeddings, incorporating surface locations inferred from depth into dense 3D grids~\cite{sscnet}, while tri-plane representations have been proposed to avoid computational burden of dense volumes~\cite{Triformer}.
Beyond fixed grids, point-based formulations have been investigated, where methods initialize dense 3D random points and refine them through iterative updates~\cite{gaussianformer,embodiedocc,opus}. Sparsity has also been leveraged: some methods construct dense 3D volumes and then discard empty voxels to obtain sparse voxel representations, which are subsequently processed using sparse convolutions~\cite{sparseocc} or transformers~\cite{voxformer,octreeocc}.

\section{Methods}\label{sec:method}

\begin{figure*}[t]
  \centering
  \includegraphics[width=0.95\textwidth]{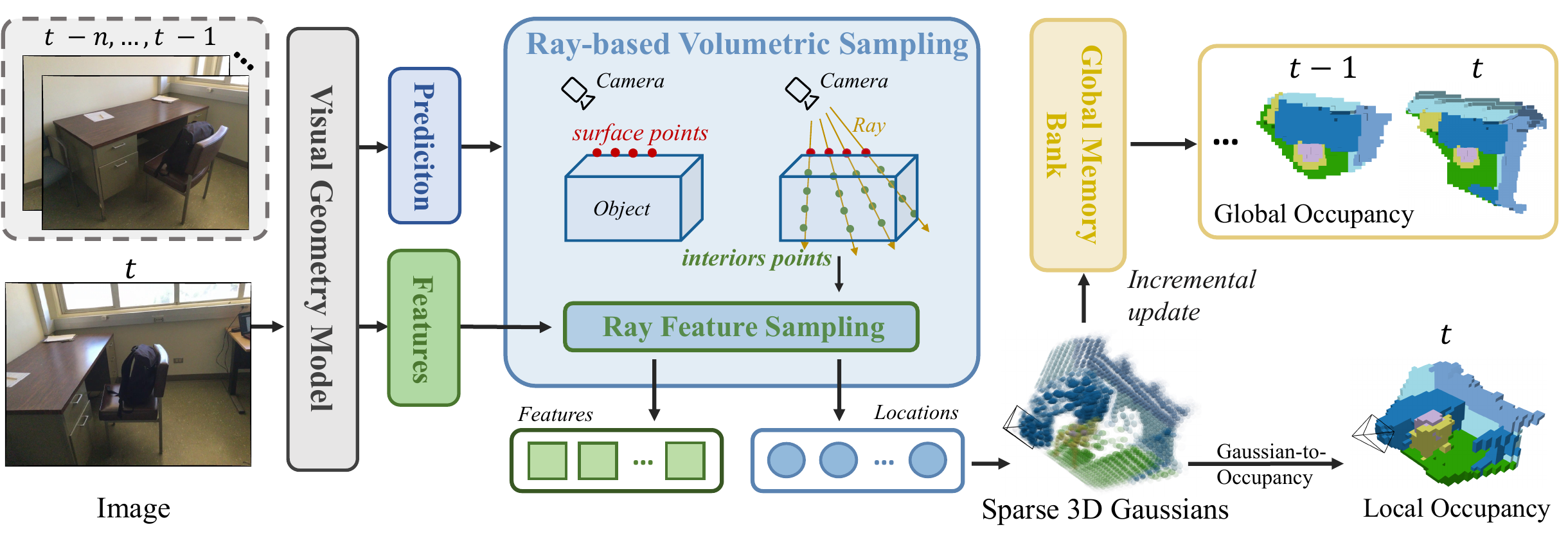}
  \caption{\textbf{Overview of \modelname.} Given an input RGB image, visual geometry priors (GPs) predicts surface points and extracts 3D-aware features. These surface points guide ray-based volumetric sampling to estimate interior points, which serve as Gaussian centers. The extracted features are combined with learnable embeddings to predict Gaussian attributes, and the resulting primitives are splatted to infer occupancy probabilistically. Monocular predictions are incrementally integrated into a global memory bank, enabling coherent large-scale occupancy construction.}
  \vspace{-5mm}
  \label{fig:framwork}
\end{figure*}

In this section, we first review the preliminaries of occupancy prediction and Gaussian splatting in~\Cref{sec:occ_gs_overview}. We then present our framework, which leverages generalizable visual geometry priors (GPs) to predict Gaussians and infer occupancy.
As illustrated in~\Cref{fig:framwork}, given an input image, visual GPs extract features and generate 3D predictions.
To address the limitation that GPs predict only surface points, which is insufficient for occupancy estimation as it requires modeling object interiors, we propose a \textit{Ray-based Volumetric Sampling} scheme (\Cref{sec:ray}). This strategy extends surface points into interior points by sampling along camera rays. Each sampled point, together with its associated features, is then used to predict Gaussian attributes.
Next, we apply an opacity-based pruning strategy to further reduce the number of Gaussians, and splat the remaining sparse Gaussians into occupancy using a probabilistic formulation~\cite{gaussianformer2} (\Cref{sec:gs2occ}). Finally, to handle embodied scenarios with streaming video inputs, we introduce a simple yet effective incremental update strategy (\Cref{sec:online}).

\subsection{Preliminaries}\label{sec:occ_gs_overview}

\noindent\textbf{Occupancy Prediction.}  
We aim to address the problem of monocular 3D occupancy prediction. Given a single RGB image $\mathbf{I}$, the goal is to predict a voxel-wise occupancy map with semantic labels $\mathbf{O} \in \mathbb{R}^{X \times Y \times Z \times N_c}$, where $X$, $Y$, and $Z$ denote the spatial resolution of the 3D scene, and $N_c$ is the number of semantic labels.  

\noindent\textbf{Gaussian Splatting.}  
A 3D scene can be compactly represented by semantic Gaussian primitives $\mathbf{G} = \{\mathcal{G}_i\}_{i=1}^P$, where each primitive $\mathcal{G}_i$ describes a local region centered at its mean $\mu_i$, and is parameterized by scale $s_i$, rotation $r_i$, opacity $a_i$, and semantic feature $c_i$. This representation is both continuous and efficient, supporting differentiable rendering and compact encoding of scene geometry and semantics.

\noindent\textbf{Gaussian-to-Occupancy.}  
Occupancy can be inferred from Gaussian primitives via Gaussian-to-voxel splatting, which aggregates the contributions of neighboring Gaussians around a voxel $p$~\cite{embodiedocc,gaussianformer,gaussianformer2}. Formally, we have:
\begin{equation}\label{eq:gs_occ}
    \hat{o}(p; \mathbf{G}) = \sum_{i \in \mathcal{N}(p)} g_i(p; \mu_i, s_i, r_i, a_i, c_i)
\end{equation}
where $\mathcal{N}(p)$ denotes the set of Gaussians that influence voxel $p$.

\subsection{Ray-based Volumetric Sampling}\label{sec:ray}

Foundation models~\cite{depthanythingv2,vggt,fast3r,point3r}, such as DepthAnything and VGGT, provide strong geometric priors (GPs) by predicting depth or point maps that capture only the \textit{visible surfaces}. However, occupancy prediction requires reasoning about both surfaces and \textit{volumetric interiors}, which is critical for embodied AI tasks such as navigation and manipulation. Surface-only reasoning fails to capture the inherent thickness of real-world objects.

To address this limitation, we propose a \textit{ray-based volumetric sampling} strategy. The core idea is to extend the predicted surface depth values along the corresponding camera rays, thereby approximating the interior volumes of objects.
Formally, given a single RGB image $\mathbf{I}\in \mathbb{R}^{H\times W \times 3}$, we extract image features $\mathbf{F}\in \mathbb{R}^{H\times W \times C}$ using a pretrained GPs to exploit the learned 3D prior.
Since depth is predicted at full resolution (which is computationally costly), we instead use intermediate features before the depth prediction layer, apply downsampling, and regress depth using a lightweight MLP:
\begin{equation}
    \mathbf{F} = \text{GPs}(\mathbf{I}),
    \quad \mathbf{F}^{\frac{1}{4}} = \text{DownSample}(\mathbf{F}), 
    \quad \mathbf{d} = \text{MLP}(\mathbf{F}^{\frac{1}{4}})
\end{equation}
where $\mathbf{F}^{\frac{1}{4}} \in \mathbb{R}^{\tfrac{H}{4} \times \tfrac{W}{4} \times C}$ denotes the downsampled feature map with spatial resolution reduced by a factor of 4, and $\mathbf{d} \in \mathbb{R}^{\tfrac{H}{4} \times \tfrac{W}{4}}$ is the predicted depth map.
Given pixel coordinate $(u,v)$ in image space, we can calculate the normalized camera ray direction $\mathbf{r}_{(u,v)}$ using camera intrinsics:
\begin{equation}
    x = \frac{u - c_x}{f_x}, \quad 
    y = \frac{v - c_y}{f_y}, \quad 
    \mathbf{r}_{(u,v)} = \frac{[\,x,\,y,\,1\,]^\top}{\sqrt{x^2 + y^2 + 1}}
\end{equation}
where $(c_x, c_y)$ are the principal point offsets, $(f_x, f_y)$ are the focal lengths, and $\mathbf{r}_{(u,v)}$ denotes the normalized camera ray direction at $(u,v)$.
To model volumetric thickness, we sample $K$ points along the ray beyond the surface point
$\mathbf{x}_{\text{surf}}$. Specifically, for pixel $(u,v)$:
\begin{equation}
    \mathbf{x}^{\text{surf}}_{(u,v)} = \mathbf{d}_{(u,v)} \cdot \mathbf{r}_{(u,v)}, \quad
    \mathbf{x}_{(u,v,k)} = \big(\mathbf{d}_{(u,v)} + \delta_k\big)\,\mathbf{r}_{(u,v)},
\end{equation}
where $k = 1,\dots,K$, $\mathbf{d}_{(u,v)}$ is the depth at $(u,v)$, and $\delta_k$ denotes the offset for the $k$-th sample, defined as
\begin{equation}\label{eq:learnable_scale}
    \{\delta_k\}_{k=1}^K = \text{linspace}(0,1,K) \cdot \text{scale}(\cdot)
\end{equation}
Here, $\text{scale}(\cdot)$ is dynamically predicted to adapt to varying object sizes.
To predict Gaussian attributes for each sampled point, we first extract its features from the image feature map $\mathbf{F}^{\tfrac{1}{4}}$. To facilitate this, we introduce a learnable embedding matrix $\mathbf{E} \in \mathbb{R}^{K \times C}$.
The point-wise features are obtained via broadcast addition:
\begin{equation} 
    \mathbf{\hat{F}}^{\tfrac{1}{4}} = \mathbf{F}^{\tfrac{1}{4}} \oplus \mathbf{E}, \quad 
    \mathbf{\hat{F}}^{\tfrac{1}{4}} \in \mathbb{R}^{\tfrac{H}{4} \times \tfrac{W}{4} \times K \times C}
\end{equation}
where $\oplus$ denotes broadcast addition and $C$ is the feature dimension. Finally, another MLP is applied to predict Gaussian attributes as follows, where $i = 1,\dots,\tfrac{H}{4}\times \tfrac{W}{4} \times K$:
\begin{equation}
\begin{aligned}
    \{\mathcal{G}_i\} &= \{s_i,r_i,a_i,c_i\}
        = \operatorname{MLP}(\mathbf{\hat{F}}^{\tfrac{1}{4}}).
\end{aligned}
\end{equation}

By augmenting surface-based predictions with ray-based volumetric sampling, our framework generates a richer set of 3D points, producing more faithful occupancy representations. This design overcomes the limitations of surface-only models like DepthAnything or VGGT and eliminates the need for dense 3D anchors~\cite{embodiedocc} or lifting 2D features into full 3D volumes~\cite{ISO}, as required by prior approaches.

\subsection{From Sparse Gaussians to Occupancy}\label{sec:gs2occ}

\begin{figure}
\vspace{-3mm}
  \centering
  \includegraphics[width=0.43\textwidth]{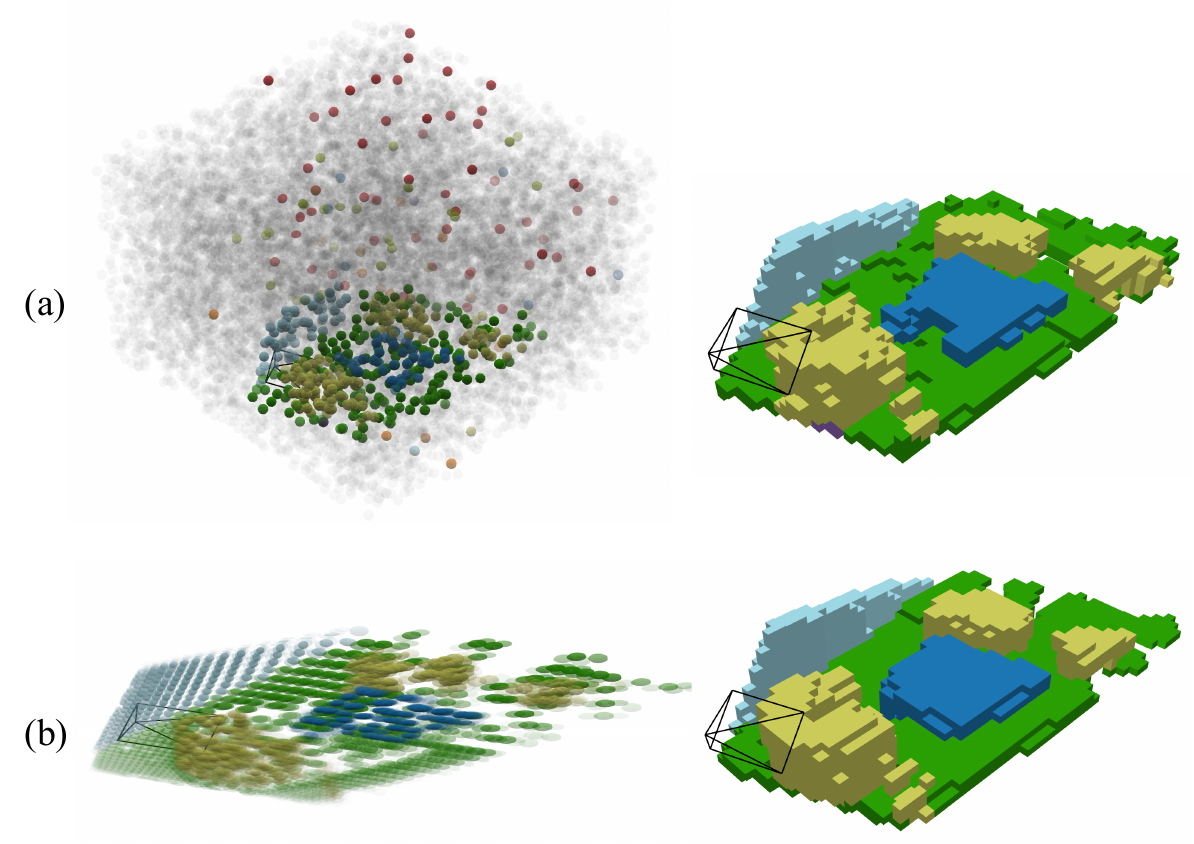}
  \caption{\textbf{Comparison of Gaussian representations.} 
(a) EmbodiedOcc, where gray indicates Gaussians predicted as empty. A substantial portion of primitives are placed in empty space, resulting in an inefficient representation. 
(b) Our method is much more compact where Gaussians are concentrated in occupied regions.}
  \label{fig:gs_compare}
\end{figure}
To cover the 3D space, prior work~\cite{embodiedocc} initializes a dense set of predefined 3D Gaussian anchors and classifies each as ``occupied'' or ``non-occupied'' to construct occupancy maps. Since most voxels in real-world scenes are empty, this strategy allocates a large fraction of primitives to non-occupied space, resulting in redundancy and inefficiency, as illustrated in~\Cref{fig:gs_compare}(a). %
In contrast, our approach leverages ray-based volumetric sampling (\Cref{sec:ray}), which naturally places Gaussians on and within objects, yielding sparse yet expressive distributions, shown in~\Cref{fig:gs_compare}(b).
To infer occupancy from these sparse Gaussians, we adopt the probabilistic Gaussian superposition formulation from GaussianFormer2~\cite{gaussianformer2}. In this formulation, regions without nearby Gaussians or far from any primitive, are naturally classified as empty voxels. Specifically, for a query point $p$, the contribution of a Gaussian $\mathcal{G}_i$ is:
\begin{equation}
o(p; \mathcal{G}_i) = \exp\left(-\tfrac{1}{2}(p - \mu_i)^\top \Sigma_i^{-1} (p - \mu_i)\right)
\end{equation}
where $\mu_i$ is the mean of the $i$-th Gaussian and $\Sigma_i$ is its covariance matrix, derived from the predicted rotation $r_i$ and scale $s_i$. $o(p; \mathcal{G}_i)$ denotes the occupancy probability at location $p$ induced by Gaussian $\mathcal{G}_i$, which smoothly decays to zero as $p$ moves away from $\mu_i$.
Importantly, the superposition of these kernels defines a continuous \emph{volumetric} occupancy field, and the learnable radial scale in~\Cref{eq:learnable_scale} controls the spatial support of each primitive along the ray. Under voxel-wise occupancy supervision, the model implicitly learns to expand or shrink the effective scales of nearby Gaussians, aligning the induced field with the
actual object extent and leaving empty regions unoccupied.

To further improve efficiency, we prune primitives with negligible opacity. Specifically, we discard all Gaussians with opacity $a_i < \tau$ (default $\tau = 0.01$), ensuring that only informative primitives contribute to the final occupancy aggregation.

\subsection{Incremental Sparse Gaussians Update}\label{sec:online}

For embodied agents that gradually perceive a scene through progressive exploration, it is important to extend monocular predictions to streaming video.
Unlike our sparse, adaptively placed Gaussians, EmbodiedOcc~\cite{embodiedocc} predefined a dense set of Gaussian anchors in advance and iteratively updates all of them, a design that does not naturally accommodate our sparse approach.
To this end, we design a training-free post-processing strategy that incrementally updates Gaussians for sequential video inputs in such scenarios.

We maintain a global Gaussian memory bank $\mathcal{M} = \{\mathcal{G}^t\}_{t=1}^M$, which accumulates Gaussians from frames $1$ through $M$. For each incoming image frame $\mathbf{I}_t$ at timestep $t$, our monocular pipeline predicts a set of Gaussians $\mathbf{G}^t = \{\mathcal{G}^{t}_i\}_{i=1}^{P_t}$ where $P_t$ denotes the total number of Gaussians predicted at this timestep.
Using camera pose, we transform all predicted centers and rotations from local camera space into a unified world coordinate system before any fusion. Each frame therefore contributes a partial sparse set of adaptively placed Gaussians in global space. 
To update the memory bank, for each $\mathcal{G}_j$ in the memory $\mathcal{M}$, we search for Gaussian neighbors $\mathcal{N}(\mathcal{G}_j) \subseteq \mathbf{G}^t$ within a spatial radius $\epsilon$ from the center of $\mathcal{G}_j$. If neighbors $\mathcal{N}(\mathcal{G}_j)$ are found, we fuse them by weighted averaging:
\begin{equation}
    \theta_i \leftarrow 
    \frac{\gamma \, p_i \theta_i + (1-\gamma) \sum_{\mathcal{G}_j \in \mathcal{N}(\mathcal{G}_i)} p_j \theta_j}
         {\gamma \, p_i + (1-\gamma) \sum_{\mathcal{G}_j \in \mathcal{N}(\mathcal{G}_i)} p_j}
\end{equation}
where $\theta \in \{\mu, \Sigma, a, c\}$ denotes the mean, covariance, opacity, and semantic feature, respectively, $p$ is top-1 class confidence, and $\gamma \in (0,1)$ controls the temporal weighting between memory and new Gaussians. We set $\gamma < 0.5$ so that newer Gaussians receive higher weight.
If no neighbors are found, the new Gaussians $\mathbf{G}^t$ are directly inserted into $\mathcal{M}$.
This incremental strategy integrates temporal information without retraining, while naturally incorporating uncertainty-aware fusion and temporal weighting for robustness.

\subsection{Training Losses}

We optimize our model with a composite objective that balances classification, segmentation, and geometric supervision. Specifically, we combine focal loss \(L_{\mathrm{focal}}\), Lovász-Softmax loss \(L_{\mathrm{lov}}\), and scene-class affinity losses \(L^{\mathrm{geo}}_{\mathrm{scal}}\) and \(L^{\mathrm{sem}}_{\mathrm{scal}}\), following EmbodiedOcc~\cite{embodiedocc}, to guide voxel-wise occupancy prediction. 
Unlike prior approaches~\cite{embodiedocc}, which rely on an external pretrained depth estimator, we add a Huber loss \(L_{\mathrm{depth}}\) directly on the predicted depth to enable end-to-end optimization of the entire pipeline, which strengthens geometric consistency between depth and occupancy, and eliminates the overhead of depth pretraining.
\begin{equation}\label{eq:loss}
\begin{split}
    \mathcal{L} & =\; L_{\mathrm{focal}}\bigl(Y_{\mathrm{mono}}^{\mathrm{fov}},\, Y_{\mathrm{gt}}^{\mathrm{fov}}\bigr)
     + L_{\mathrm{lov}}\bigl(Y_{\mathrm{mono}}^{\mathrm{fov}},\, Y_{\mathrm{gt}}^{\mathrm{fov}}\bigr) \\
    & + L^{\mathrm{geo}}_{\mathrm{scal}}\bigl(Y_{\mathrm{mono}}^{\mathrm{fov}},\, Y_{\mathrm{gt}}^{\mathrm{fov}}\bigr)
     + L^{\mathrm{sem}}_{\mathrm{scal}}\bigl(Y_{\mathrm{mono}}^{\mathrm{fov}},\, Y_{\mathrm{gt}}^{\mathrm{fov}}\bigr)
     + L_{\mathrm{depth}}
\end{split}
\end{equation}

\section{Experiments}\label{sec:exp}

\begin{table*}[h] \small
    \caption{\textbf{Monocular prediction performance on the Occ-ScanNet dataset.}}
    \vspace{-7mm}
    \small
    \setlength{\tabcolsep}{0.008\textwidth}
    \captionsetup{font=scriptsize}
    \begin{center}
    \resizebox{0.98\linewidth}{!}{
    \begin{tabular}{l|c|c c c c c c c c c c c|c}
        \toprule
        Method
        & {IoU}
        & \rotatebox{90}{\parbox{1.5cm}{\textcolor{ceiling}{$\blacksquare$} ceiling}} 
        & \rotatebox{90}{\textcolor{floor}{$\blacksquare$} floor}
        & \rotatebox{90}{\textcolor{wall}{$\blacksquare$} wall} 
        & \rotatebox{90}{\textcolor{window}{$\blacksquare$} window} 
        & \rotatebox{90}{\textcolor{chair}{$\blacksquare$} chair} 
        & \rotatebox{90}{\textcolor{bed}{$\blacksquare$} bed} 
        & \rotatebox{90}{\textcolor{sofa}{$\blacksquare$} sofa} 
        & \rotatebox{90}{\textcolor{table}{$\blacksquare$} table} 
        & \rotatebox{90}{\textcolor{tvs}{$\blacksquare$} tvs} 
        & \rotatebox{90}{\textcolor{furniture}{$\blacksquare$} furniture} 
        & \rotatebox{90}{\textcolor{objects}{$\blacksquare$} objects} 
        & mIoU\\
        \midrule
        TPVFormer~\cite{Triformer} & 33.39 & 6.96 & 32.97 & 14.41 & 9.10 & 24.01 & 41.49 & 45.44 & 28.61 & 10.66 & 35.37 & 25.31 & 24.94 \\
        GaussianFormer~\cite{gaussianformer} & 40.91 & 20.70 & 42.00 & 23.40 & 17.40 & 27.0 & 44.30 & 44.80 & 32.70 & 15.30 & 36.70 & 25.00 & 29.93 \\
        MonoScene~\cite{monoscene} & 41.60 & 15.17 & 44.71 & 22.41 & 12.55 & 26.11 & 27.03 & 35.91 & 28.32 & 6.57 & 32.16 & 19.84 & 24.62 \\
        ISO~\cite{ISO} & 42.16 & 19.88 & 41.88 & 22.37 & 16.98 & 29.09 & 42.43 & 42.00 & 29.60 & 10.62 & 36.36 & 24.61 & 28.71 \\
        Surroundocc~\cite{surroundocc} & 42.52 & 18.90 & 49.30 & 24.80 & 18.00 & 26.80 & 42.00 & 44.10 & 32.90 & 18.60 & 36.80 & 26.90 & 30.83 \\
        EmbodiedOcc~\cite{embodiedocc} & 53.55 & 39.60 & 50.40 & 41.40 & 31.70 & 40.90 & 55.00 & 61.40 & 44.00 & 36.10 & 53.90 & 42.20 & 45.15 \\
        EmbodiedOcc++~\cite{embodiedocc++} & 54.90 & 36.40 & 53.10 & 41.80 & 34.40 & 42.90 & 57.30 & 64.10 & 45.20 & 34.80 & 54.20 & 44.10 & 46.20 \\
        \midrule
        Ours-DPT & \underline{56.96} & \underline{51.42} & \underline{50.35} & \underline{46.97} & \underline{41.84} & \underline{46.98} & \underline{60.39} & \underline{66.16} & \underline{50.51} & \underline{47.97} & \underline{58.88} & \underline{49.23} & \underline{51.88} \\
        Ours-VGGT & \textbf{63.14} & \textbf{51.67} & \textbf{59.93} & \textbf{52.07} & \textbf{46.44} & \textbf{51.35} & \textbf{64.45} & \textbf{69.47} & \textbf{54.30} & \textbf{51.76} & \textbf{63.29} & \textbf{53.36} & \textbf{56.19} \\
        \bottomrule
    \end{tabular}
    }
    \end{center}
    \vspace{-5mm}
    \label{tab:main_mono}
 \end{table*}

\begin{table*}[h] 	\small
    \caption{\textbf{Embodied prediction performance on the EmbodiedOcc-ScanNet dataset.}}
    \vspace{-5mm}
    \setlength{\tabcolsep}{0.007\textwidth}
    \captionsetup{font=scriptsize}
    \begin{center}
    \resizebox{0.98\linewidth}{!}{
    \begin{tabular}{l|c|c c c c c c c c c c c|c}
        \toprule
        Method
        & {IoU}
        & \rotatebox{90}{\parbox{1.5cm}{\textcolor{ceiling}{$\blacksquare$} ceiling}} 
        & \rotatebox{90}{\textcolor{floor}{$\blacksquare$} floor}
        & \rotatebox{90}{\textcolor{wall}{$\blacksquare$} wall} 
        & \rotatebox{90}{\textcolor{window}{$\blacksquare$} window} 
        & \rotatebox{90}{\textcolor{chair}{$\blacksquare$} chair} 
        & \rotatebox{90}{\textcolor{bed}{$\blacksquare$} bed} 
        & \rotatebox{90}{\textcolor{sofa}{$\blacksquare$} sofa} 
        & \rotatebox{90}{\textcolor{table}{$\blacksquare$} table} 
        & \rotatebox{90}{\textcolor{tvs}{$\blacksquare$} tvs} 
        & \rotatebox{90}{\textcolor{furniture}{$\blacksquare$} furniture} 
        & \rotatebox{90}{\textcolor{objects}{$\blacksquare$} objects} 
        & mIoU\\
        \midrule
        TPVFormer~\cite{Triformer} & 35.88 & 1.62 & 30.54 & 12.03 & 13.22 & 35.47 & 51.39 & 49.79 & 25.63 & 3.60 & 43.15 & 16.23 & 25.70 \\
        SurroundOcc~\cite{surroundocc} & 37.04 & 12.70 & 31.80 & 22.50 & 22.00 & 29.90 & 44.70 & 36.50 & 24.60 & 11.50 & 34.40 & 18.20 & 26.27 \\
        GaussianFormer~\cite{gaussianformer} & 38.02 & 17.00 & 33.60 & 21.50 & 21.70 & 29.40 & 47.80 & 37.10 & 24.30 & 15.50 & 36.20 & 16.80 & 27.36 \\
        SplicingOcc~\cite{embodiedocc} & 49.01 & 31.60 & 38.80 & 35.50 & 36.30 & 47.10 & 54.50 & 57.20 & 34.40 & 32.50 & 51.20 & 29.10 & 40.74 \\
        EmbodiedOcc~\cite{embodiedocc} & 51.52 & 22.70 & 44.60 & 37.40 & 38.00 & 50.10 & 56.70 & 59.70 & 35.40 & 38.40 & 52.00 & 32.90 & 42.53 \\
        EmbodiedOcc++~\cite{embodiedocc++} & 52.20 & 27.90 & 43.90 & 38.70 & 40.60 & 49.00 & 57.90 & 59.20 & 36.80 & 37.80 & 53.50 & 34.10 & 43.60 \\
        \midrule
        Ours-DPT & \underline{56.39} & \underline{40.80} & \underline{48.78} & \underline{45.62} & \underline{43.26} & \underline{50.08} & \underline{63.97} & \underline{67.72} & \underline{48.36} & \underline{48.77} & \underline{60.46} & \underline{45.63} & \underline{51.22} \\
        Ours-VGGT & \textbf{61.41} & \textbf{42.61} & \textbf{51.35} & \textbf{51.49} & \textbf{48.72} & \textbf{54.32} & \textbf{67.91} & \textbf{70.73} & \textbf{52.94} & \textbf{54.75} & \textbf{64.76} & \textbf{49.67 }& \textbf{55.39} \\
        \bottomrule
    \end{tabular}
    }
    \end{center}
    \vspace{-3mm}
    \label{tab:main_online}
 \end{table*}

\subsection{Datasets and Metrics}

\noindent\textbf{Occ-ScanNet}~\cite{ISO} is a large-scale benchmark for monocular indoor occupancy prediction, containing 45,755 training samples and 19,764 testing samples. The dataset covers diverse scenes and viewpoints, and provides voxelized frames in $60 \times 60 \times 36$ grids, corresponding to a $4.8 \text{m} \times 4.8 \text{m} \times 2.88 \text{m}$ volume in front of the camera. Each voxel is annotated with 12 semantic classes, including 11 valid categories (ceiling, floor, wall, window, chair, bed, sofa, table, TV, furniture, objects) and one class for empty space.  

\noindent\textbf{EmbodiedOcc-ScanNet}~\cite{embodiedocc} is a reorganized version of Occ-ScanNet, consisting of 537 training scenes and 137 validation scenes. Each scene contains 30 posed frames, and the global occupancy resolution of a scene is defined as $\frac{l_x \times l_y \times l_z}{(0.08\text{m})^3}$
where $l_x \times l_y \times l_z$ denotes the spatial range of the scene in world coordinates.

\noindent\textbf{Evaluation Metrics.}
We follow prior work and adopt mIoU and IoU as evaluation metrics. Specifically, for {Occ-ScanNet}~\cite{ISO,embodiedocc}, we compute IoU between predictions and ground truth within the camera frustum of each frame. For {EmbodiedOcc-ScanNet}~\cite{embodiedocc}, we evaluate global occupancy by computing IoU at the scene level, where the entire reconstructed scene is considered.

\subsection{Implementation Details}

For all our models we adopt the same training strategy. We employ the AdamW optimizer~\cite{adamw} with a weight decay of 0.01. The learning rate is linearly warmed up during the first 1000 iterations to a maximum value of $2\times10^{-4}$ and then decayed following a cosine schedule. The model is trained for 10 epochs with a total batch size of 8 on 4 NVIDIA A800 GPUs. Input images are resized such that the longer side is 518 pixels, following the setting in VGGT~\cite{vggt}. We apply gradient clipping with a maximum norm of 1.0. Unless otherwise specified, we set $K=16$ for ray-based volumetric sampling, and $\tau=0.01$ for opacity-based pruning.

\subsection{Occupancy Prediction Results}

\noindent\textbf{Results on Occ-ScanNet.}
As shown in~\Cref{tab:main_mono}, bold numbers denote the best result and underlined numbers denote the second best. Overall, \modelname~with VGGT priors achieves 63.14 IoU and 56.19 mIoU, surpassing EmbodiedOcc and EmbodiedOcc++ by +11.04 and +9.99 mIoU, respectively.
Using DepthAnything as the geometry prior, same as~\cite{embodiedocc,ISO}, Ours-DPT model already delivers consistent improvment performance across all classes over previous method and improves mIoU from 46.20 (EmbodiedOcc++) to 51.88. This indicates that our design more effectively exploits GPs for volumetric reasoning. Furthermore, replacing DepthAnything with the stronger VGGT prior yields additional gains, showing that our framework generalizes well across different geometry priors and can fully benefit from more powerful foundation models.

\noindent\textbf{Results on EmbodiedOcc-ScanNet.} 
\Cref{tab:main_online} reports the embodied prediction performance on the EmbodiedOcc-ScanNet benchmark. 
Our VGGT-based model (Ours-VGGT) achieves 61.41 IoU and 55.39 mIoU, which is +9.2 IoU and +11.8 mIoU higher than the previous SoTA EmbodiedOcc++~\cite{embodiedocc++}.
Notably, the DepthAnything variant (Ours-DPT) already attains 56.39 IoU and 51.22 mIoU, outperforming EmbodiedOcc++ by +4.2 IoU and +7.6 mIoU, which highlights the effectiveness of our streaming fusion strategy. Leveraging the stronger VGGT prior further boosts performance, showing that \modelname~can robustly exploit different GPs to maintain coherent scene representations over time in streaming and embodied settings.

\begin{figure*}[h]
  \centering
  \includegraphics[width=0.98\textwidth]{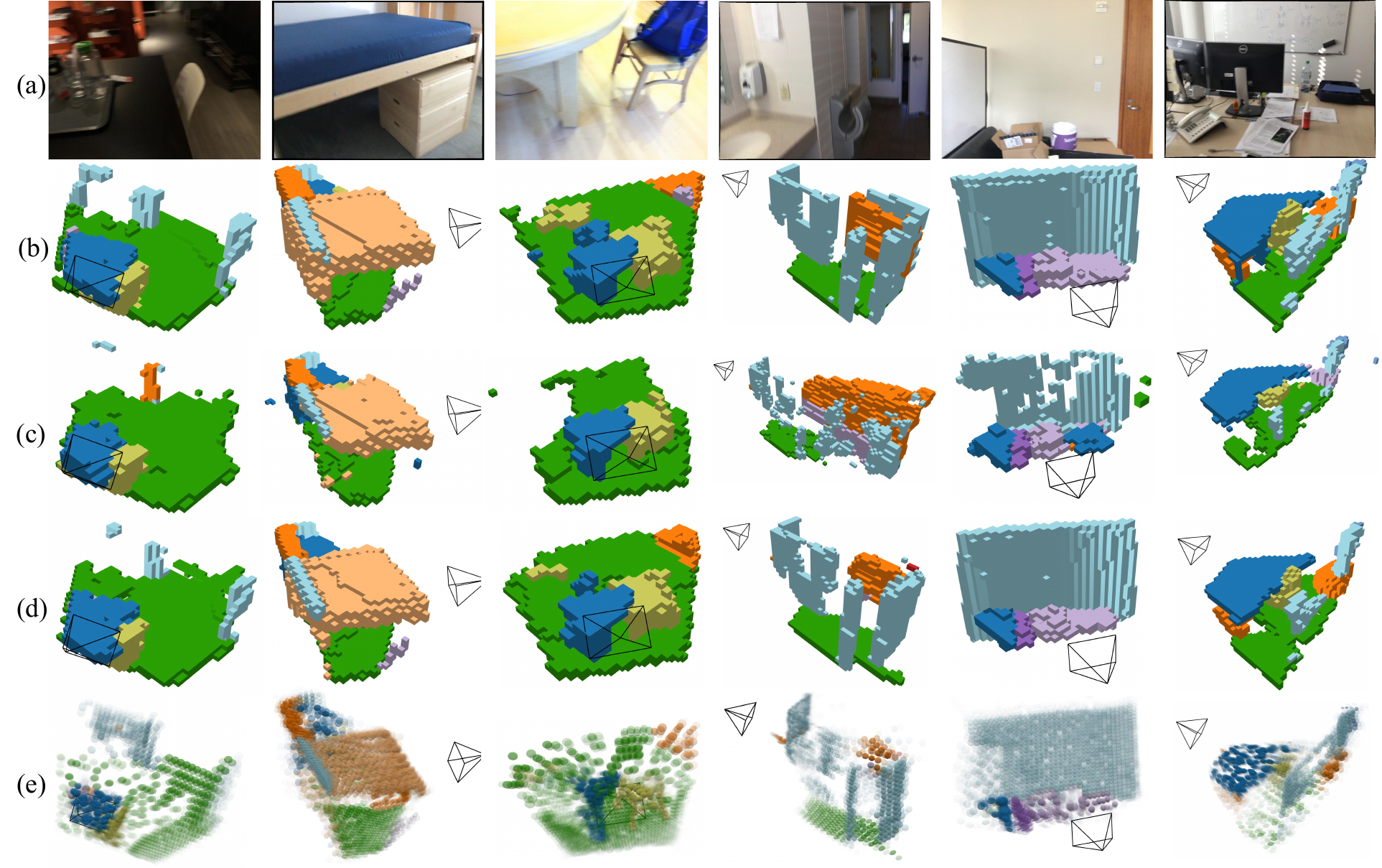}
  \vspace{-1mm}
  \caption{\textbf{Qualitative comparison on monocular occupancy prediction.}
 (a) shows the input RGB images, (b) the ground-truth occupancy, (c) the predictions of EmbodiedOcc~\cite{embodiedocc}, (d) the predictions of our method, and (e) the visualization of the Gaussian primitives predicted by our method. Compared to EmbodiedOcc, our framework produces more accurate and complete reconstructions, while the Gaussian representation provides interpretable intermediate geometry.}
  \vspace{-1mm}
  \label{fig:mono_vis}
\end{figure*}

\begin{figure*}[h]
  \centering
  \includegraphics[width=0.98\textwidth]{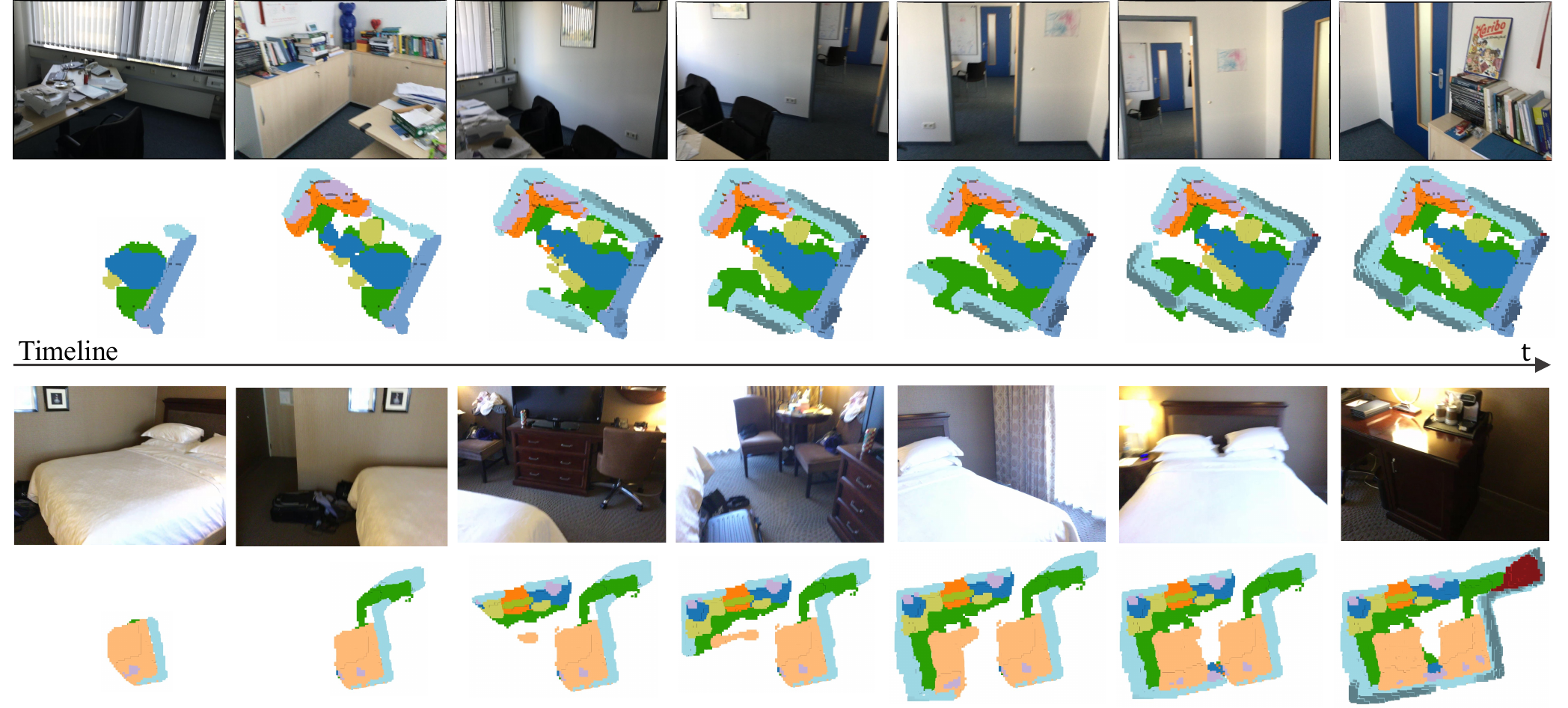}
  \vspace{-1mm}
  \caption{\textbf{Qualitative results on streaming inputs.}
  Our incremental update strategy progressively integrates information from sequential frames. The predictions become increasingly complete as more frames are observed, demonstrating the effectiveness of our streaming design.}
  \vspace{-1mm}
  \label{fig:emb_vis}
\end{figure*}

\subsection{Ablation studies}

\begin{table}[h]\small
\caption{\textbf{Comparison on the Occ-ScanNet dataset in terms of accuracy, efficiency, and model complexity.} 
DPT and VGGT refer to~\cite{depthanythingv2} and~\cite{vggt}, respectively.
All FPS values are measured on the same computer with NVIDIA A800 GPU, averaged over 1000 runs after an initial warm-up of 100 iterations.}
\vspace{-2mm}
\centering
\resizebox{0.95\linewidth}{!}{
    \begin{tabular}{lcccc}
    \toprule
    \textbf{Model} & IoU & mIoU & FPS & \#Params \\
    \midrule
    ISO~\cite{ISO} & 42.16 & 28.71 & 3.63  & 303.05M \\
    EmbodiedOcc~\cite{embodiedocc} & 53.55 & 45.15 & 10.66 & 231.45M \\
    \midrule
    Ours-DPT  & 56.96 & 51.88 & \textbf{28.22} & \textbf{97.95M} \\
    Ours-VGGT & 63.14 & 56.19 & 5.26        & 942.31M \\
    \bottomrule
    \end{tabular}
}
\label{tab:fps}
\vspace{-3mm}
\end{table}

\noindent\textbf{Comparison with Different Geometry Priors.}
To further validate our design, we instantiate our framework with the DepthAnything backbone, following ISO~\cite{ISO} and EmbodiedOcc~\cite{embodiedocc}. 
Unlike ISO, which feeds depth into a heavy 3D U-Net, or EmbodiedOcc, which stacks a complex decoder on top of the backbone, \textbf{Ours-DPT} attaches only a lightweight Gaussian head without any 3D convolutional decoder. 
As shown in~\Cref{tab:fps}, this variant already achieves stronger performance while being significantly more efficient: with the same depth prior, \textbf{Ours-DPT} improves IoU from 53.55 to 56.96 and mIoU from 45.15 to 51.88 over EmbodiedOcc, while running nearly $3\times$ faster (28.22 vs.\ 10.66 FPS) with less than half the parameters (97.95M vs.\ 231.45M).
\textbf{Ours-VGGT} further pushes accuracy to 63.14 IoU and 56.19 mIoU, while keeping the same lightweight occupancy head. 
Taken together, these results show that our novel way of leveraging geometry priors yields both higher accuracy and substantially better efficiency, making fast and reliable occupancy prediction feasible for downstream embodied applications.

\begin{table}[h]
\small
\centering
\caption{\textbf{Ablation on the number of sampled points $K$ per ray.} Increasing $K$ improves accuracy, but gains saturate beyond $K{=}16$ despite a sharp increase in the number of Gaussians.}
\vspace{-3mm}
\label{tab:abl_bins}
\begin{tabular}{c|ccc}
\toprule
\(K\) & mIoU & IoU & \#Gaussians \\
\midrule
1  & 47.88 & 53.10 &  3079 \\
2  & 52.65 & 56.35 &  2310 \\
4  & 55.28 & 60.35 &  2731 \\
8  & 55.76 & 61.67 &  3940 \\
16 & 56.19 & 63.14 &  5876 \\
32 & 56.72 & 63.84 & 20206 \\
\bottomrule
\end{tabular}
\end{table}

\begin{table}[h]
\small
\centering
\caption{\textbf{Ablation on the opacity threshold $\tau$.} Lower thresholds (e.g., $>0.01$) perform best; larger thresholds prune too many Gaussians and degrade accuracy.}
\vspace{-3mm}
\label{tab:abl_opa}
\begin{tabular}{c|ccc}
\toprule
Threshold & mIoU & IoU & \#Gaussians \\
\midrule
$>0.01$ & 56.19 & 63.14 & 5876 \\
$>0.02$ & 55.21 & 62.08 & 3509 \\
$>0.03$ & 54.59 & 61.45 & 2468 \\
$>0.04$ & 54.29 & 61.09 & 1931 \\
$>0.05$ & 54.16 & 60.84 & 1612 \\
$>0.06$ & 54.02 & 60.54 & 1398 \\
$>0.07$ & 53.82 & 60.13 & 1239 \\
$>0.08$ & 53.51 & 59.62 & 1116 \\
$>0.09$ & 53.12 & 59.01 & 1014 \\
$>0.10$ & 52.65 & 58.31 &  930 \\
\bottomrule
\end{tabular}
\end{table}

\noindent\textbf{Effect of number of sampled points $K$ along each ray.}
We study the impact of sampled number $K$ in~\Cref{tab:abl_bins}. Increasing $K$ consistently improves performance, as more samples enrich the coverage of volumetric interiors. However, the improvement saturates beyond $K=16$, where IoU and mIoU exhibit only marginal gains at $K=32$ despite a substantial increase in the number of Gaussians.

\noindent\textbf{Effect of opacity threshold $\tau$.}
We conduct ablation studies on the choice of opacity threshold $\tau$ as shown in~\Cref{tab:abl_opa}. A smaller threshold (\eg, $>0.01$) yields the best performance, achieving 56.19 mIoU and 63.14 IoU. As $\tau$ increases, performance gradually degrades due to excessive pruning.

\subsection{Qualitative Results}

We present qualitative comparisons on monocular occupancy prediction in~\Cref{fig:mono_vis}, where we compare against EmbodiedOcc~\cite{embodiedocc}. In addition,~\Cref{fig:emb_vis} illustrates two examples of our incremental update strategy on streaming inputs. More visualizations are provided in the Appendix.

\section{Conclusion}

We proposed \modelname, a novel framework that leverages visual geometry priors for fine-grained occupancy prediction. By extending surface points along camera rays into volumetric samples, our method constructs sparse Gaussian primitives that compactly capture scene geometry. We further introduced an incremental update mechanism that adapts the framework seamlessly to streaming video inputs. Extensive experiments on Occ-ScanNet and EmbodiedOcc-ScanNet show that \modelname~achieves SoTA accuracy and generalizes across different priors, providing a scalable and effective solution for embodied 3D perception. Looking ahead, we believe \modelname~offers a promising step toward integrating strong geometric priors into broader embodied AI tasks, including interactive scene understanding, navigation and manipulation.

{
    \small
    \bibliographystyle{ieeenat_fullname}
    \bibliography{main}
}

\end{document}